\documentclass[conference]{IEEEtran}
\IEEEoverridecommandlockouts
\usepackage{cite}
\usepackage{url}
\usepackage{amsmath,amssymb,amsfonts}
\usepackage{algorithmic}
\usepackage{graphicx}
\usepackage{textcomp}
\usepackage{xcolor}
\def\BibTeX{{\rm B\kern-.05em{\sc i\kern-.025em b}\kern-.08em
    T\kern-.1667em\lower.7ex\hbox{E}\kern-.125emX}}

\newcommand{\mc}{\mathcal}
    
\begin{document}
\title{Enhancing Embedding Representations of \\ 
Biomedical Data using Logic Knowledge
}

\author{\centering\IEEEauthorblockN{Michelangelo Diligenti}
\IEEEauthorblockA{\textit{DIISM, University of Siena} \\
Siena, Italy \\
michelangelo.diligenti@unisi.it}
\and
\IEEEauthorblockN{Francesco Giannini}
\IEEEauthorblockA{\textit{DIISM, University of Siena}\\
Siena, Italy \\
francesco.giannini@unisi.it}
\and
\IEEEauthorblockN{Stefano Fioravanti}
\IEEEauthorblockA{\textit{DIISM, University of Siena}\\
Siena, Italy \\
stefano.fioravanti66@gmail.com}
\and\hspace{2cm}
\and
\IEEEauthorblockN{Caterina Graziani}
\IEEEauthorblockA{\textit{DIISM, University of Siena}\\
Siena, Italy \\
caterina.graziani@student.unisi.it}
\and
\IEEEauthorblockN{Moreno Falaschi}
\IEEEauthorblockA{\textit{DIISM, University of Siena}\\
Siena, Italy \\
moreno.falaschi@unisi.it}
\and
\IEEEauthorblockN{Giuseppe Marra}
\IEEEauthorblockA{\textit{DCS, KULeuven}\\
Leuven, Belgium \\
giuseppe.marra@kuleuven.be}}

\maketitle

\begin{abstract}

Knowledge Graph Embeddings (KGE) have become a quite popular class of models specifically devised to deal with ontologies and graph structure data, as they can implicitly encode statistical dependencies between entities and relations in a latent space. KGE techniques are particularly effective for the biomedical domain, where it is quite common to deal with large knowledge graphs underlying complex interactions between biological and chemical objects.
Recently in the literature, the PharmKG dataset has been proposed as one of the most challenging knowledge graph biomedical benchmark, with hundreds of thousands of relational facts between genes, diseases and chemicals.
Despite KGEs can scale to very large relational domains, they generally fail at representing more complex relational dependencies between facts, like logic rules, which may be fundamental in complex experimental settings.
In this paper, we exploit logic rules to enhance the embedding representations of KGEs on the PharmKG dataset.
To this end, we adopt  Relational Reasoning Network (R2N), a recently proposed neural-symbolic approach showing promising results on  knowledge graph completion tasks. An R2N uses the available logic rules to build a neural architecture that reasons over KGE latent representations. In the experiments, we show that our approach is able to significantly improve  the current state-of-the-art on the PharmKG dataset.  
Finally, we provide an ablation study to experimentally compare the effect of alternative sets of rules according to different selection criteria and varying the number of considered rules.
\end{abstract}


\section{Introduction}
\label{sec:introduction}
Knowledge Graph Embeddings (KGEs) have been established as an efective class of models to process large  relational datasets by encoding entities and relations as real-valued embedding vectors  \cite{nickel2015review,dai2020survey}. By learning these embedded representations according to a known set of relational facts, described as triples of the form $(\textit{entity}, \textit{relation}, \textit{entity})$, 
KGEs are able to generalize predictions to unknown triples. However, KGEs generally struggle in capturing more complex relational dependencies among facts, like First-Order Logic (FOL) rules that can be useful to infer new triples with respect to an incomplete set of given facts.
Several variants of KGE integrating additional information, like entity types and logic rules, have been reviewed in Wang et al.\cite{wang2017knowledge}. Combining KGE with additional semantic information or with reasoning schema is currently a fundamental line of research. Some of the main advancements in this area have been collected in some recent surveys like Li et al.~\cite{li2020hybrid} and Zhang et al.~\cite{zhang2021neural}.

Among this class of approaches, Relational Reasoning Networks (R2N) \cite{marra2021relational} have been recently proposed as a neural-symbolic model integrating KGEs with neural networks. The main idea behind this model is to process the embedded representations of facts provided by a KGE as a message-passing algorithm over a logic-based factor graph, defined by the available set of ground logic rules.
In its original version, R2Ns showed promising results for knowledge graph completion tasks but it was tested only on small datasets, where it was possible to manually compile the domain knowledge. 
This paper instead considers the application of R2N to knowledge completion in a large biomedical knowledge graph, where the extraction of the domain knowledge is carried out by using automatic tools for logic rules mining.

The PharmKG dataset~\cite{zheng2021pharmkg} is a recently proposed multi-relational benchmark with hundreds of thousands of fact triples between genes, drugs and diseases, concerning approximately 7000 entities. As several relation types on the entities are considered, we claim that logic rules can constitute a valuable source of additional knowledge to establish complex interconnections among these entities.
However, given the remarkable dimension of this benchmark, successfully applying methodologies that combine KGEs and logic rules is very challenging and, according to the authors' knowledge, this is the first work accomplishing this task.
The main contributions of this paper can be summarized as follows:
\begin{itemize}
    \item[(i)] we show how processing logic knowledge with
    R2N  may significantly improve the prediction performances on a large-size knowledge graph completion task;
    \item[(ii)] we apply our methodology on the PharmKG benchmark, achieving a new state-of-the-art with a large margin;
    \item[(iii)] we show how it is possible to employ automatic rule mining techniques to enrich knowledge graph completion tasks, and we provide an in-depth ablation study to evaluate the performance of R2Ns according to different settings of the rule miner.
\end{itemize}
The paper is organized as follows: Section \ref{sec:back} introduces the basics on KGE and FOL theories, together with a description on how logic rules can be modeled to define factor graphs representing the logic structure. In Section \ref{sec:R2N}, we present Relational Reasoning Networks, which is the model we rely on to integrate KGE with FOL rules. Section \ref{sec:expsetup} describes the composition of the PharmKG dataset with some relevant statistics, and the basics of the rule mining model that was used to extract the rules. The experimental results and the ablation study are discussed in Section \ref{sec:exp}, while Section \ref{sec:relwork} reviews the literature on related approaches. Finally, Section \ref{sec:conc} summarizes the main results and discusses future followups of this work.

\section{Background}
\label{sec:back}

\subsection{Knowledge Graph Embeddings}

In recent years, Knowledge Graphs (KGs) have emerged as a common approach for the representation of human knowledge~\cite{wang2017knowledge,dai2020survey}. 
Knowledge graphs consist of facts in form of triples formed by two entities and a relation. A KG can be seen as a graph, where each entity is a node and each relation establishes an edge between the entities in the fact. For example, the fact $\textit{Friends}(\textit{Alice},\textit{Bob})$ links the entities $\textit{Alice}$ and $\textit{Bob}$ with an edge labeled by the relation $\textit{Friends}$.
KGs are generally incomplete and Knowledge Graph Embeddings (KGE) are a powerful approach for populating KGs by mapping entities and relations to latent representations, which generalize the assignments to unknown facts. For instance in our previous example, $\textit{Alice},\textit{Bob}$ and $\textit{Friends}$ are represented as real-valued vectors $e_A$, $e_B$, $W_F$, and KGE methods learn these embeddings by defining scoring functions that are trained to match the supervisions. 
All KGE methods first reconstruct a fact representation from the embeddings of the entities and relation, and then compute a score from this representation via a scoring function. Given a fact $R(e_1,e_2)$, in the following we provide some examples of popular KGEs.
\begin{itemize}
    \item \emph{TransE}~\cite{bordes2013translating} models relations as translation operations on the embeddings of the entities, where the internal atom representation $e_1 + W_R - e_2$ is scored as $1 / (1 + \|e_1 + W_R - e_2\|)$.
    \item \emph{DistMult}~\cite{yang2015embedding} computes the dot product among entity and relation embeddings $<\!e_1, e_2, W_R\!>$, which corresponds to an internal atom representation given by the Hadamard product $a=e_1 \cdot e_2 \cdot W_R$, and $\sigma(\sum_i a_i)$ as score function, being $\sigma$ the sigmoid function.
    \item \emph{ComplEx}~\cite{trouillon2016complex} uses the Hermitian dot product over complex embeddings to also model asymmetric relations: $Re(<e_1, e_2, W_R>)$.
    \item \emph{Neural Tensor Networks} \cite{socher2013reasoning} model asymmetric relations via a bilinear formulation: $u_R^T f(e_1^T W_R e_2 + V_R [e_1, e_2] + b)$, with $V_R, u_R, b$ additional weight vectors, and $f=\tanh$.
\end{itemize}

\begin{figure}[t]
    \centering
    \includegraphics[width=0.48\textwidth]{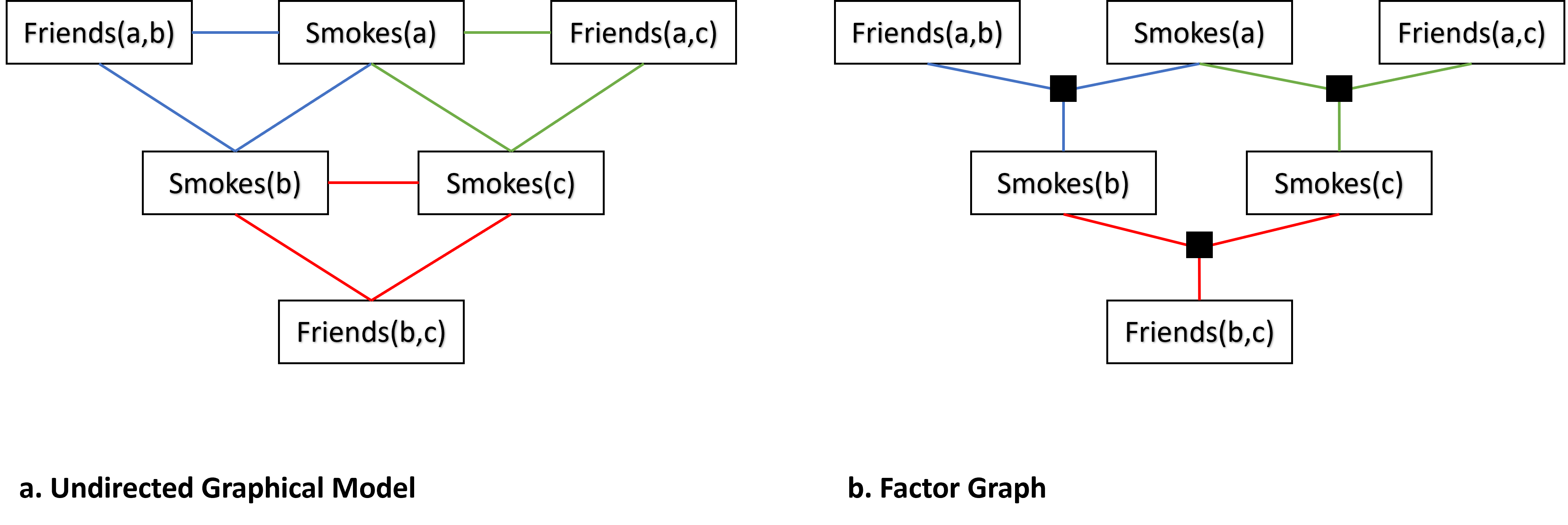}
    \caption{Given the logic rule $\forall x\forall y\ \textit{Friends}(x,y) \wedge \textit{Smokes}(x) \rightarrow \textit{Smokes}(y)$, grounded for the constants $\{a, b, c\}$, we represent in  {\bf a.} its associated undirected graphical model and in {\bf b.} the factor graph associated to this formula. In the picture, each clique is highlighted with a different colour.} 
    \label{fig:graphical_model_to_transformer}
\end{figure}

\subsection{First-Order Logic}

A possible way of considering knowledge graphs is as a set of (function-free) First-Order Logic (FOL) atomic formulas. However, a FOL language is generally much more expressive and consists of a set of variables $x_i$, constants $c_i$, predicates $P_i$, logic connectives, and quantifiers.
In particular, constants and (binary) predicates trivially denote entities and relations as defined in a KG, but FOL predicates can also have arbitrary arity.
Any $n$-ary predicate $P$ represents a property that may hold true or false for an $n$-ary tuple of constants, for $n\geq 1$. For instance $\textit{Alice}$ and $\textit{Bob}$ can be represented by constants $c_a,c_b$, while
$\textit{Friends}(x,y)$ denotes a binary predicate. However, other examples like $\textit{Smokes}(x)$ or $\textit{SonOf}(x,y,z)$, expressing that $x$ smokes or that $x$ is the son of $y$ and $z$, are modeled as well in FOL as a unary and a ternary predicate, respectively. 
Given an $n$-ary predicate $P$ and  $t_1,\ldots,t_n$ variables/constants, $P(t_1,\ldots,t_n)$ is called an atom, and more specifically a \textit{ground atom} if each $t_i$ is a constant $c_i$. FOL formulas are then defined by compounding atoms with connectives and quantifiers in the standard way \cite{smullyan1995first}. A set of FOL formulas is said a logic \textit{theory}. A formula $\varphi$ where no variables occur is called a {ground formula}. Given a FOL formula $\forall x_1\ldots\forall x_n\ \varphi(x_1,\ldots,x_n)$, the process of \textit{grounding} consists in removing all the quantifiers and replacing the variables $x_1,\ldots,x_n$ with some constants $c_1,\ldots,c_n$, thus obtaining a ground formula $\varphi(c_1,\ldots,c_n)$.

\begin{figure*}[ht]
    \centering
\includegraphics[width=0.75\textwidth]{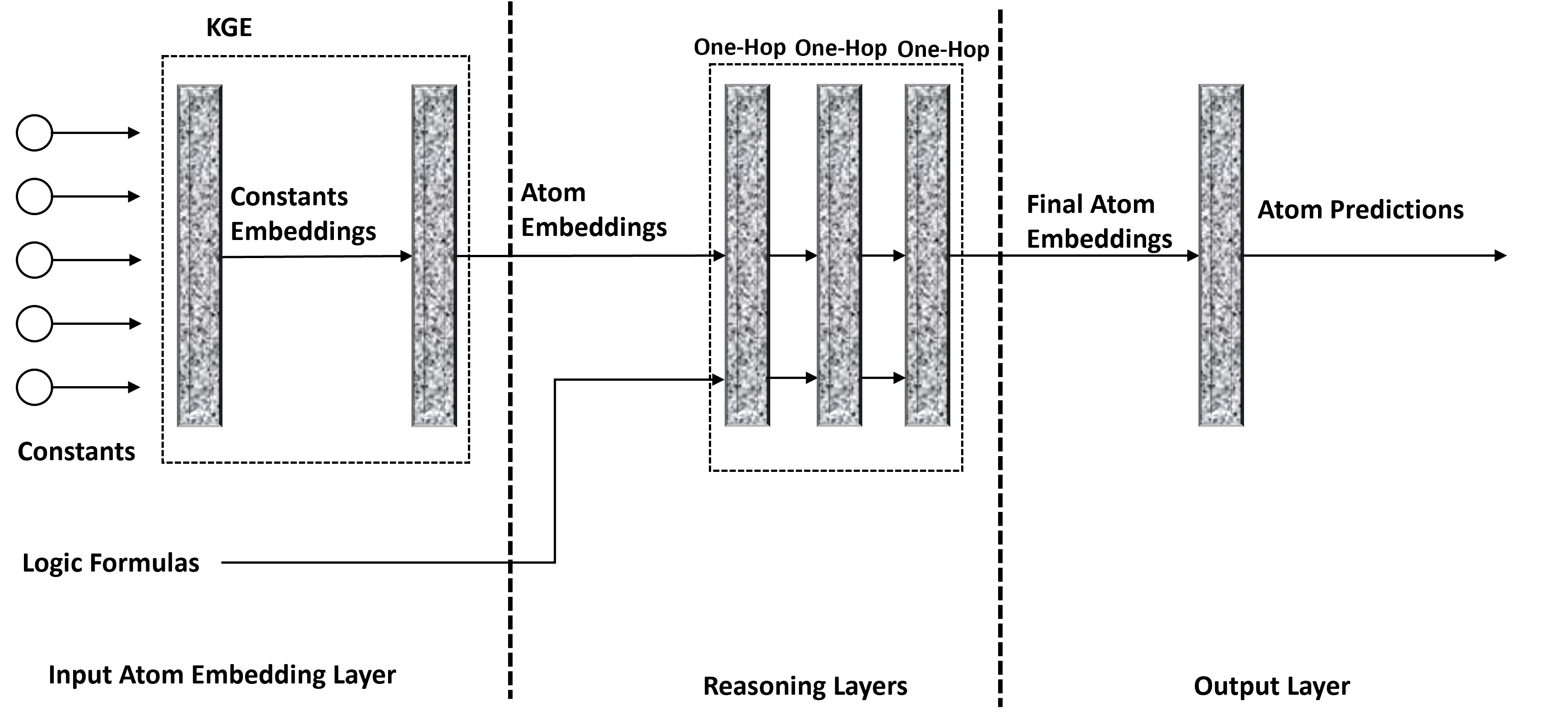}
    \caption{Overview of an R2N with a KGE providing input embeddings and multiple reasoning layers allowing multi-hop reasoning. 
    }
    \label{fig:overall_r2n}
\end{figure*}

\subsection{From FOL Theories to Factor Graphs}
\label{sec:backFG}

Statistical Relational Learning (SRL) methods \cite{de2010statistical,khosravi2010survey}, like well-known Markov Logic Networks \cite{richardson2006markov}, use FOL theories as a template to build an undirected graphical model by considering all the possible groundings of the formulas. More specifically, the ground FOL theory is mapped to a factor graph \cite{loeliger2004introduction}, where ground atoms are mapped to nodes and there exists an edge between two nodes if the two nodes appear in the same ground formula. According to this mapping, there is a clique among the nodes that correspond to the set of ground atoms co-occurring in the same ground formula.
The factor graph can be defined by replacing one factor for each maximal clique in the graphical model, and connecting each node in the model to the factors for the cliques it is associated to.
Reasoning can be performed on this graph by exploiting standard algorithms~\cite{bishop2006pattern}, but exact inference is generally intractable and approximation techniques are required to perform both parameter learning and probabilistic reasoning within the model.
For example, the undirected graphical model corresponding to $\forall x\forall y\ \textit{Friends}(x,y)\wedge \textit{Smoke}(x)\rightarrow \textit{Smoke}(y)$ with constant domain $D=\{a,b,c\}$ is represented in Figure \ref{fig:graphical_model_to_transformer}-a, where each clique is highlighted with a different color. The associated factor graph is shown in Figure~\ref{fig:graphical_model_to_transformer}-b. 

Relational Reasoning Networks consider the flattened relational structure of a factor graph for a given grounded FOL theory. The network structure is used to encode the reasoning process by manipulating the embedding representations of the atoms, taking the initial representation provided by a KGE.

\section{Foundations of Relational\\ Reasoning Networks}
\label{sec:R2N}

Knowledge Graph Embeddings are an effective methodology to work with large collections of relational data. However, they have been shown 
to be quite limited in terms of reasoning capabilities, especially when logic 
knowledge is available, 
and this motivate the study of several approaches applying neural or logic inference on KGE outputs
\cite{li2020hybrid,zhang2021neural}. 
In this section, we recall the basics of  Relational Reasoning Networks \cite{marra2021relational}, a recently introduced model that can get advantage of KGE representations and some available logic knowledge in a fully integrated fashion. For a detailed description of the model, we refer the interested reader to the original paper \cite{marra2021relational}.

\subsection{Overall View}

A Relational Reasoning Network (R2N) is a neural architecture, whose underlying structure mimics logic reasoning, where new logic facts are computed from known ones by applying inference rules. However, an R2N works exclusively at a latent level, where logic facts (i.e. atoms) are represented as embeddings and logic knowledge is used to update such embeddings in a completely neural architecture. 
This approach presents the advantage of leaving the network itself learning the embedding representations of the atoms that better model the available set of known facts. In this way the logic rules can be used in a more versatile way, thus providing the network the chance to freely modify the atom embeddings according to the logic structures or ignoring them, in case the rules are not providing useful additional information to solve the learning task.
Another advantage of R2Ns is that they are agnostic to the input representation of the atoms, which can be feature vectors or embeddings. For example, the input atoms take the form of KGE embeddings in this paper.




The structure of a Relational Reasoning Network is defined according to three kinds of layers, combining a KG with a logic-based factor graph. As shown in Figure \ref{fig:overall_r2n}, the \textit{Input Atom Embedding Layer} takes a set of logic facts (i.e. triples in the KG) and computes a first latent representation using a standard KGE. 
This representation is projected multiple times by a stack of  \textit{Reasoning Layers}, which enhance the input representations with the relational information coming from the logic rules, in the shapes of a logic factor graph. Finally, an  \textit{Output Layer} provides the final atom predictions.

\subsection{Neural Reasoning on Embedding Representations}
A reasoning layer of an R2N is a neural architecture that takes as input: (i) a latent representation of the logic atoms and (ii) a logic theory as a factor graph (see Section \ref{sec:backFG}). Then, it returns a transformed representation of the logic atoms, taking into account the relations with other atoms. 

The reasoning process takes the shape of a message-passing algorithm among atom nodes and factor nodes. While this mapping of reasoning to a message-passing algorithm is well-known in the SRL field, R2N exploits messages with latent semantics, allowing a much higher learning capacity. This is what provides the model the flexibility of learning when and on which atoms to apply the grounded rules.
Intuitively, each factor node representation is computed by a neural network that aggregates the messages of the atoms that appear in the corresponding ground rule. Similarly, the new atom representations are computed by aggregating the messages of the factors/rules in which they appear. 

In particular, at each reasoning layer $l$, atom nodes send a message $m_{a \rightarrow f}$ to their neighbors factor nodes:

\begin{equation*}
    m^{l}_{a \rightarrow f} = x^{l-1}_a
\end{equation*}
where $x^{l-1}_a$ is the embedding of the atom $a$ at layer $l-1$. Notice that $x^0_a$ is the embedding computed by the KGE.

Then, the factor node representation is computed as:

\begin{equation*}
    x^{l}_{f} = MLP^f(m^{l}_{a_1 \rightarrow f}, ..., m^{l}_{a_m \rightarrow f})
\end{equation*}
where $\{a_1, ..., a_m\} \equiv ne(f)$ are the  neighbors atom nodes of $f$ in the factor graph.

Similarly, we can compute messages and updated representations of atom nodes. In particular:

\begin{equation*}
m^{l}_{f \rightarrow a} = MLP^a_{j,i}(x^{l}_{f}) 
\end{equation*}
\begin{equation*}
x^{l}_{a} = x^0_{a} + \sum_{f \in ne(a)} m^{l}_{f \rightarrow a} 
\end{equation*}
where $MLP^a_{j,i}$ indicates a specific neural network for each rule $r_j$ and for each position $i$ that the atom $a$ takes in rule $r_j$. We notice that $x^{l}_{a}$ has been defined as a residual with respect to the KGE embedding $x_a^0$ to facilitate the network to ignore the messages coming from the logic knowledge when not useful to improve the atom predictions.
\\

\noindent \textbf{Multi-Hop Reasoning.} The reasoning process can be repeated multiple times by providing the newly computed representation of the atoms, i.e. $x^{l}_{a}$, to a new layer $l+1$. This represents a fundamental property, akin to multi-hop reasoning in logic-based systems. Indeed, we expect the model to discover new facts by applying the rules to the input facts. Therefore, we would like to exploit such new knowledge to trigger more rules in subsequent steps of reasoning. 
\\

\noindent \textbf{Output Computation.} Finally, an output layer takes the final reasoning representations at layer $l=L$ and outputs the ground atom prediction $y_a$ using a fully connected layer with a single output and sigmoidal activation:
\begin{align*}
y_{a} &  = MLP_o(x^L_{a})
\end{align*}

\section{Experimental Setup}
\label{sec:expsetup}

This section summarizes the experimental background that has been used in Section \ref{sec:exp} to evaluate our approach. In particular, Section \ref{sec:pharmkg} describes the PharmKG dataset together with a few statistics we calculated on its relational structure. Section \ref{sec:AMIE} recalls the basic properties of AMIE, a rule mining model we used to extract the rules and instantiate R2N. Section~\ref{sec:mining_rules_on_pharmkg} discusses some examples of these extracted rules, also from the biological plausibility point of view.

\subsection{PharmKG Dataset}
\label{sec:pharmkg}

PharmKG \cite{zheng2021pharmkg} is a multi-relational, attributed, biomedical Knowledge Graph, composed of 500.360 individual interconnections between genes, drugs, and diseases, with 28 relation types over a vocabulary of 7262 disambiguated entities. 
This dataset is publicly available in the Pykeen library\footnote{\url{https://pykeen.readthedocs.io/en/stable/_modules/pykeen/datasets/pharmkg.html#PharmKG}} and it is detailed as follows.
PharmKG is composed of triples of the form (\textit{entity}, \textit{relation}, \textit{entity}) where the entities belong to three different domains according to the following distributions: 4732 genes, 1496 chemicals, and 1034 diseases. 
Some examples of the considered relations are the predicate $An$, meaning ``ancestors of disease", $Te$ representing ``possible therapeutic effect", or the ``pharmacokinetics" $K$ that relates genes and chemicals.
We refer to the original paper \cite{zheng2021pharmkg} for further details on the semantics of the available relations in this dataset.

\begin{table*}[t]
\centering
\caption{Frequency of the relations relative to their domain and percentage of triples satisfying the reflexive, symmetric, and transitive properties. For short we will use the shortcut $\mc{G}:=\textit{Genes}$, $\mc{D}:=\textit{Diseases}$ and $\mc{C}:=\textit{Chemicals}$.}
\begin{tabular}{c|c|c|c|c|c}
\multicolumn{1}{c|}{\textbf{Domain}} &
\multicolumn{1}{|c|}{\textbf{Relation}} &
\multicolumn{1}{|c|}{\textbf{Frequency}} &
\multicolumn{1}{|c|}{\textbf{Reflexive}} &
\multicolumn{1}{|c|}{\textbf{Symmetric}} &
\multicolumn{1}{|c}{\textbf{Transitive}} \\
\hline
$\mc{G} \times \mc{G}$ & Ra & 19.24 \% & 17.44 \% & 45.09 \% & 26.31 \%\\
 & GG & 42.13 \% & 0.07 \% & -- & 25.01 \%\\
 & Rg & 22.16 \% & -- & -- & --\\
 & Q & 16.47 \% & 34.86 \% & 46.82 \% & 29.12 \% \\\hline
$\mc{G} \times \mc{D}$ & U & 26.49 \% & -- & -- & --\\
 & P & 58.87 \%  & -- & -- & --\\
  & D & 4.10 \%  & -- & -- & --\\
  & Te & 10.54 \%  & -- & -- & --\\\hline
$\mc{G} \times \mc{C}$ &  K & 66.90 \% & -- & -- & -- \\
 & O & 21.16 \% & -- & -- & -- \\
 & Z & 11.94 \%   & -- & -- & --\\\hline
$\mc{D} \times \mc{D}$ & An & 94.94 \% & 53.22 \% & 21.01 \% & 83.29 \% \\
 & As & 5.06 \% & 15.38 \% & 50.0 \% & 22.86 \% \\\hline
$\mc{D} \times \mc{G}$ & ML & 92.73 \% & -- & -- & -- \\
 & X & 7.27 \% & -- & -- & -- \\\hline
 $\mc{C} \times \mc{G}$ & N & 91.03 \% & -- & -- & -- \\
 & A & 8.97 \% & -- & -- & -- \\\hline
 $\mc{C} \times \mc{D}$ & T & 61.39 \% & -- & -- & -- \\
& Sa & 21.01 \% & -- & -- & -- \\
& J & 4.96 \% & -- & -- & -- \\
& Pr & 9.35 \% & -- & -- & -- \\
& C & 3.29 \% & -- & -- & -- \\\hline
$(\mc{G} \cup \mc{C})\times \mc{G}$ & E & 67.56 \% & 11.65 \% & 29.5 \% & 19.53 \% \\
  & B & 32.44 \%  & 13.18 \% & 37.26 \% & 12.96 \%\\\hline
$\mc{C} \times \mc{C}$ & CC & 100 \% & -- & -- & 30.93 \%\\\hline
$\mc{D} \times \mc{C}$ & Mp & 100 \% & -- & -- & -- \\\hline
$(\mc{G}  \cup \mc{C})\times(\mc{G} \cup \mc{C} \cup \mc{D})$ & I & 100 \% & -- & 9.15 \% & 2.2 \% \\\hline
$(\mc{G} \cup \mc{C} \cup \mc{D})\times(\mc{G} \cup \mc{C} \cup \mc{D})$ & Iw & 100 \% & 3.7 \% & 3.39 \% & 15.5 \%
\end{tabular}
\label{tab:freq by domain}
\end{table*}

In order to point out few insights on how the considered relations depend on each other, 
we computed some statistics on the dataset. In particular, we calculated relations' relative frequency by domain and in the whole dataset, and we inspected whether some classical properties of relations are satisfied, such as reflexivity, symmetricity and transitivity. For clarity, we recall here the definitions of these properties for a certain relation $R$ with entities $x,y,z$:
\[
\begin{array}{ll}
     \textit{reflexivity} &  \forall x\ R(x,x) \\
     \textit{symmetricity} & \forall x\forall y\ R(x,y)\rightarrow R(y,x)\\
     \textit{transitivity} & \forall x\forall y\forall z\ R(x,y)\wedge R(y,z)\rightarrow R(x,z)
\end{array}
\]
\begin{figure}
   \centering
   \includegraphics[scale=.5]{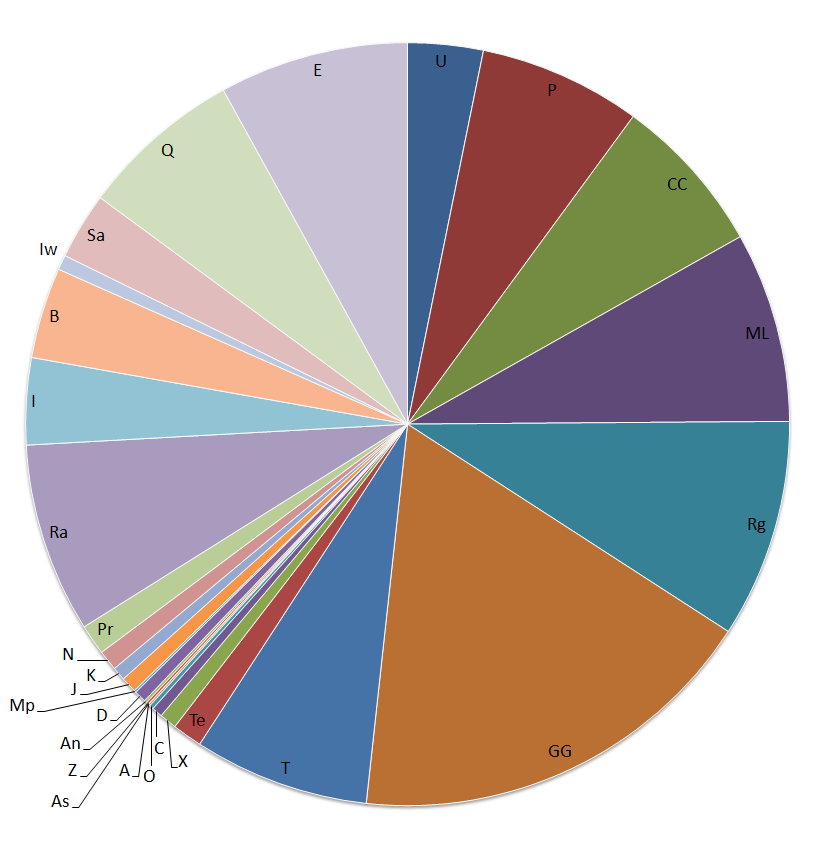}
   \caption{Relative frequency of the relations in the whole dataset.}
   \label{fig:rel_freq}
\end{figure}
The relative frequency of relations, as shown in Figure \ref{fig:rel_freq}, visualizes the distribution of edge types, and most frequent types are generally easier to learn by standard KGEs.
Table \ref{tab:freq by domain} reports the frequency of each relation in the dataset recomputed per domain and the percentage of triples that satisfy the reflexive, symmetric, and transitive rules. We use the symbol ``$-$" to denote that the considered relation is neither reflexive nor symmetric, while it is trivially transitive as there are no triples of entities in the domain linked by the same relation.
Please note that all relations are not strictly reflexive, symmetric, or transitive: this may be due to the actual incompleteness of the training data, or due to the intrinsic probabilistic nature of these regularities. Therefore, it is important to employ a reasoner like R2N, which has the flexibility of learning the manifolds where some given knowledge applies.



\subsection{Rule Extraction and Selection}
\label{sec:AMIE}
Logic rules, holding among the atoms of a KG, can help to deduce and add missing knowledge to an incomplete set of known facts. Such rules can be either manually devised when working on a restrained problem or, more commonly in large knowledge graphs, can be automatically extracted.
Indeed, many neuro-symbolic approaches employ an external rule miner to extract the rules, like for example~\cite{nakashole2012query,richardson2006markov}.
In this paper, logic rules have been mined from the PharmKG dataset using a popular framework known as \textit{Association rule Mining under Incomplete Evidence in ontological knowledge bases} (AMIE) \cite{galarraga2013amie}, which is also a common choice for several others models integrating KGE and logic rules, like \cite{guo2018knowledge,cheng2021uniker}.
AMIE mines the rules, together with a set of \textit{confidence} scores, computed according to different criteria. In Section \ref{sec:abl} we performed an ablation study on three confidence metrics to understand which rules are the best candidates for integration with R2N. These confidence metrics, all rely on the definition of rule support.

\noindent\textbf{Rule Support. }
The support of a rule with head $R(x,y)$ is defined as the number of distinct pairs of subjects and objects in the head of all instantiations that appear in the KG:
\begin{equation*}
   supp(\Vec{B}\Rightarrow R(x,y)) = \#(x,y) : \exists z_1, \dots z_m : \Vec{B} \Rightarrow R(x,y)
\end{equation*}
where $\Vec{B} = B_1 \wedge \dots \wedge B_n$ and $z_1, \dots z_m$ are the variables of the rule apart from $x$ and $y$.
In the following, we report the rule confidence scores provided by AMIE.
\begin{itemize}
    \item \textbf{Head Coverage:} this score is the proportion of pairs from the head relation that are covered by the predictions of the rule (a rule recall metric):
\begin{equation*}
    hc(\Vec{B} \Rightarrow R(x,y)):= \frac{ 
supp(\Vec{B}\Rightarrow R(x,y))}{\#(x',y'):R(x',y')}
\end{equation*}

    \item\textbf{Standard Confidence:} under a closed world-assumption, where all facts that are not in the KG are negative evidence, the standard confidence of a rule is the ratio of its predictions that are in the KG (a rule precision metric):
\begin{equation*}
    conf(\Vec{B}\Rightarrow R(x,y)) = \frac{supp(\Vec{B} \Rightarrow R(x,y))}{\#(x,y): \exists z_1 \dots z_m : \Vec{B}}
\end{equation*}
Standard confidence blinds the distinction between “false” and “unknown”, thus implementing a closed world assumption.

    \item\textbf{PCA Confidence:} the \textit{partial completeness assumption} (PCA) states that if the database knows some $R$-attribute of some variable $x$, then it knows all $R$-attributes of $x$. Under the PCA, the confidence is normalized by the set of known true facts, instead of by the entire set of facts, together with the facts assumed to be false:
\begin{equation*}
    pca(\Vec{B} \!\Rightarrow\! R(x,y)) \!=\! \frac{supp(\Vec{B} \!\Rightarrow\! R(x,y))}{\#(x,y) \! : \! \exists z_1, \dots, z_m, y' \!:\! \Vec{B} \!\land\! R(x,y')}
\end{equation*}
\end{itemize}

\subsection{Mining Rules on PharmKG}
\label{sec:mining_rules_on_pharmkg}
\begin{table}[t]
    \centering
    \caption{Examples of rules used in the experimental evaluation. In the arguments of the rules we use the symbols $c \in \mc{C}$, $d \in \mc{D}$, $g \in \mc{G}$.}
    \label{tab:std_rules}
    \begin{tabular}{l|c}
    Rule  & Standard Confidence\\
    \hline
$An(d,d_1)\wedge Iw(d,d_1) \rightarrow An(d_1,d) $ & 0.909\\
$Iw(d,g)\wedge ML(d,g) \rightarrow Iw(g,d) $ & 0.833\\
$Q(g,g_1) \wedge Rg(g,g_1) \rightarrow Q(g_1,g)   $ & 0.825 \\
$Iw(d,g)\wedge U(g,d) \rightarrow Iw(g,d)  $ & 0.818 \\
$Q(g,g_1) \wedge Rg(g,g_1) \rightarrow Rg(g_1,g)  $ & 0.800 \\
$B(g,g_1) \wedge Q(g,g_1) \rightarrow Q(g_1,g)    $ & 0.793 \\
$E(g,g_1) \wedge Q(g,g_1) \rightarrow Q(g_1,g)    $ & 0.793 \\
$B(g,g_1) \wedge Rg(g,g_1) \rightarrow Rg(g_1,g)  $ & 0.793 \\
$GG(g,g_1)\wedge Q(g,g_1) \rightarrow Q(g_1,g)   $ & 0.792 \\
$Q(g,g_1) \rightarrow Q(g_1,g)            $ & 0.786 \\
$Ra(g,g_1)\wedge Rg(g,g_1) \rightarrow Rg(g_1,g) $ & 0.783  
    \end{tabular}
\end{table}
The AMIE framework was used to extract a set of logic rules, in form of Horn Clauses, that can then be used to define the structure of an R2N.
The $10$ rules with the highest AMIE standard confidence are listed in Table~\ref{tab:std_rules}.

A closer inspection of the rules shows that some of them have a clear biological meaning. In the following, we report some interesting examples.
\begin{itemize}
    \item $An(d_1,d_2) \wedge An(d_2,d_3) \Rightarrow An(d_1,d_3)$, this rule expresses the transitivity property for the ``\textit{ancestor of disease}" relation $An$.
\item $D(g,d) \Rightarrow Te(g,d) $, where $D$ is the relation ``\textit{drug target}" connecting a gene to a disease, and $Te$ relation indicates a ``\textit{possible therapeutic effect}" on the pathology.
This logic rule states that if a gene is a drug target for a certain disease, then it can have a therapeutic effect on the disease.
 \item $P(g,d) \wedge X(d,g) \Rightarrow ML(d,g)$, where the $P,X,ML$ relations indicate a ``\textit{pathogenesis link}", ``\textit{over-expression}", and ``\textit{diagnostic biomarker}", respectively.
 The extracted rule states that whenever the relations $P$ and $X$ hold between a gene $g$ and a disease $d$, then the gene is a diagnostic biomarker for the pathology.
\end{itemize}

\section{Experimental Results}
\label{sec:exp} 
This section reports an experimental evaluation addressing the following questions: 
\begin{itemize}
\item how R2N prediction accuracy compares against what can be obtained using its main competitors on the PharmKG dataset;\footnote{The software package, data, and scripts to reproduce the results can be downloaded from \url{http://UrlHiddenForDoubleBlindReview}.}

\item which is the effect of the number of considered rules and their selection criteria on the R2N results.
\end{itemize}

\subsection{Competitive Evaluation}
PharmKG is based on a complex interconnected multi-relational setting, where logic rules representing expert domain knowledge are not immediately available. 
The  methodology proposed in this paper has been evaluated against different models, like standard KGE (such as TransE, TransR, RESCAL, ComplEx, and DistMult), neural-based methods (such as ConvE, ConvKB, and RGCN), and the previous state-of-the-art model on PharmKG (HRGAT). 
In the experiments, we achieved the best performances using an R2N with two stacked reasoning layers, where all MLPs used in the reasoning layers have one single layer and ReLu activation functions. The R2Ns employed the DistMult KGE as input atom embedding layer.
The embedding sizes for the ground atoms and ground formulas have been fixed to 250 and 25, respectively. 
The rules exploited to define the R2N structure have been extracted using AMIE, as explained in Section~\ref{sec:mining_rules_on_pharmkg}, considering the $100$ rules with highest AMIE standard confidence.

In order to reduce the number of considered cliques in the factor graph, we used the commonly employed heuristic of considering only the ground formulas whose premises are verified in the training set, like similarly done by other neuro-symbolic methods like ExpressGNN~\cite{zhang2020efficient}, UniKER~\cite{cheng2021uniker}.
The output predictions have been trained using the binary cross-entropy loss, 
using the positive examples from the dataset plus a sample of $2$ random corruptions for each positive example. We used the Adam optimizer with a learning rate of $10^{-2}$ to pre-train the KGE models for 250 epochs and then we fine tuned the R2N for another 30 epochs using a learning rate equal to $10^{-3}$. 
Table~\ref{tab:results} reports the experimental results of R2N against its competitors.

\begin{table}[t]
    \centering
    \caption{Knowledge Graph Completion metrics of the proposed methods and the main competitors (results from \cite{zheng2021pharmkg}) on the PharmKG dataset.}
    \label{tab:results}
    \begin{tabular}{l|c ccc}
        \textbf{Model} & MRR & Hits@1 & Hits@3 & Hits@10\\
        \hline
        TransE \cite{bordes2013translating} & 0.091 & 0.034 & 0.092 &0.198\\
        TransR \cite{lin2015learning} & 0.075 & 0.030 & 0.071 &0.155\\
        RESCAL \cite{nickel2011three} & 0.064 &0.023& 0.057 &0.122\\
        ComplEx \cite{trouillon2016complex} & 0.107 & 0.046 & 0.110 & 0.225\\
        DistMult \cite{yang2015embedding} & 0.063 &0.024& 0.058 & 0.133\\
        ConvE \cite{dettmers2018convolutional} & 0.086 &0.038& 0.058 & 0.169\\
        ConvKB \cite{nguyen2017novel} & 0.106 &0.052& 0.107 &0.209 \\
        RGCN \cite{schlichtkrull2018modeling} & 0.067 &0.027& 0.062 &0.139\\
        HRGAT \cite{zheng2021pharmkg}  & 0.154 &0.074& 0.172 & 0.315\\
        \textbf{R2N} & {\bf 0.215} & {\bf 0.145} & {\bf 0.234} & {\bf 0.342}
    \end{tabular}
\end{table}
As we can see from the results, our model significantly outperforms all the other methods on all the evaluation metrics. Thus showing that exploiting additional logic knowledge with R2N, even if possibly noisy and automatically extracted by a rule miner, can make the difference in solving this task.

\subsection{Ablation Study}
\label{sec:abl} 

An ablation study was performed to investigate the effect of the rules extracted by AMIE and fed to the R2N. In particular, the first $N$ (with $5 \le N \le 100$) most relevant rules, ordered according to the three confidence metrics described in \ref{sec:AMIE}, have been considered for each experiment.

\begin{figure}[t]
    \centering
    \includegraphics[width=0.48\textwidth]{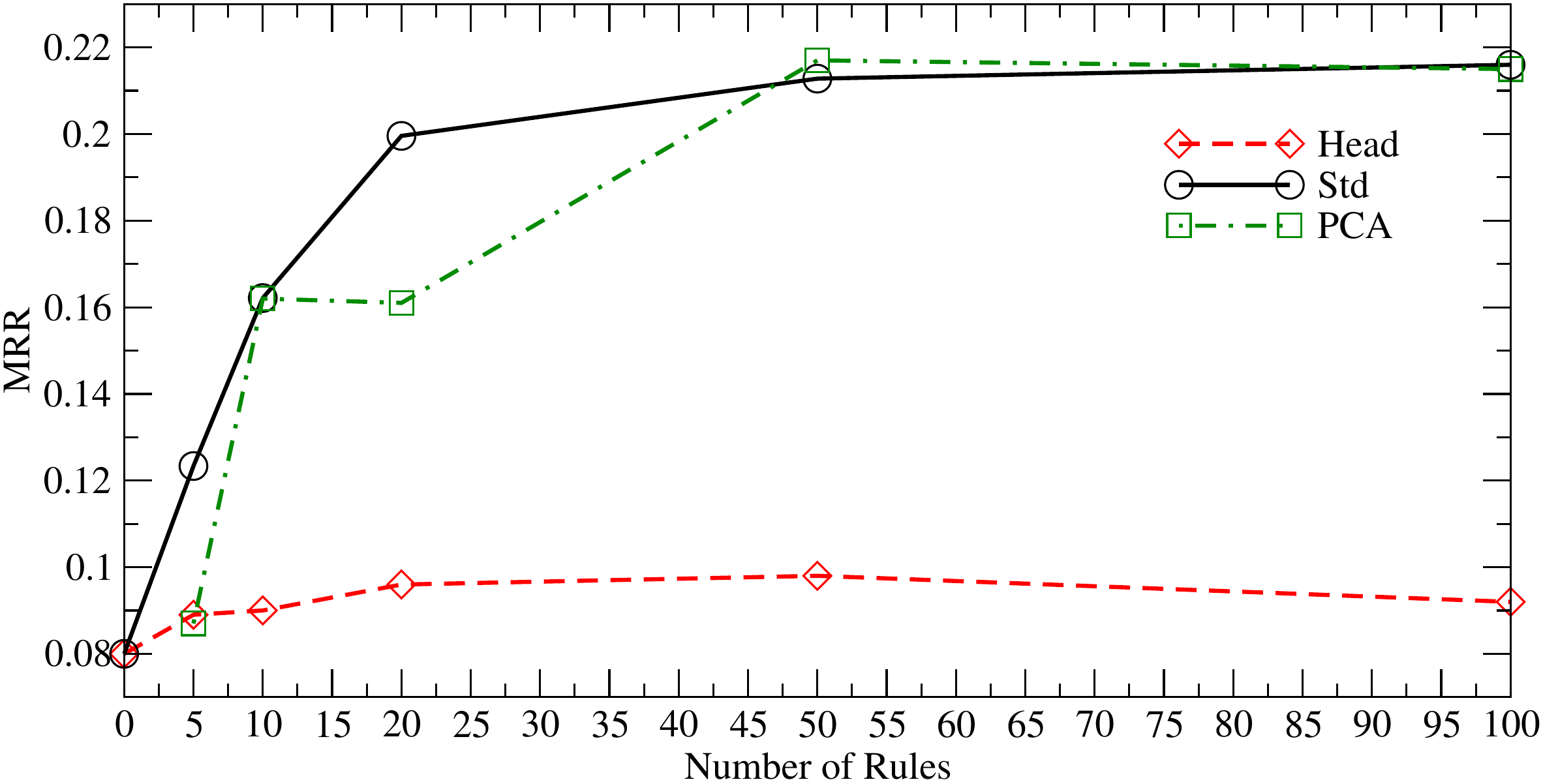}
    \caption{MRR results varying the sorting criterium and the number of selected rules.}
    \label{fig:ablation_results}
\end{figure}

Figure \ref{fig:ablation_results} reports the scores of the MRR on the test set with respect to different types and number of rules.
As we can notice from the results, the criteria to select the rules and how many rules are considered, have a strong influence on the R2N performances. As expected, an increase in the number of rules improves the breadth of the reasoning process and increases the number of predicates, which can be affected by the reasoner, thus providing better final results. We found that the rules with high standard or PCA confidence perform similarly, whereas selecting rules in terms of head coverage is not very effective. Head coverage rules provide only a small gain over the baseline, which does not grow with the increase of the rule number. This result is justified by the intuition that the head coverage of a rule measures its recall, e.g. if there are alternative explanations for a fact, but it not representing its precision, e.g. how many times it leads to a wrong conclusion.
On the other hand, precision-based metrics like standard and PCA confidence, tend to select rules that are very targeted and easier to exploit.


\section{Related Work}
\label{sec:relwork}

\subsection{Biomedical Knowledge Graphs}
KGs are a meaningful way of representing knowledge in the biomedical domain. Important examples include knowledge bases on the effect of environmental factors on human health (CTD,  \cite{davis2019comparative}), the impact of human genetic variation on drug response (PharmGKB,  \cite{hewett2002pharmgkb}), human genes and genetic phenotypes (OMIM,  \cite{hamosh2000online}) and drugs and drug targets (DrugBank,  \cite{wishart2008drugbank}).
Belleau et al. \cite{belleau2008bio2rdf} were the first to publish a major biomedical knowledge graph (KG) work, using semantic web technologies to convert publicly available bioinformatics databases into RDF formats, extending them with 
a significant number of metadata relations.
Percha et al. \cite{percha2018global} compiled a rough KG, known as the Global Network of Biomedical Relationships (GNBR) from large-scale biomedical literature with unsupervised techniques and used it to generate drug re-purposing hypotheses. 

Recently, \cite{walsh2020biokg} introduced PharmKG, a large knowledge graph built from publicly available databases and text-mined knowledge bases. The authors proposed an extensive evaluation benchmark on this dataset using knowledge graph embedding techniques. The benchmark included both domain-agnostic and domain-specific (i.e. biomedical-oriented) architectures.  
The dataset is also accompanied by pre-trained entity embeddings constructed from both graph methods and text-based methods.

\subsection{Logic-based KGE}

Logic rules provide higher-order semantics among the considered relations. Numerous rule-based KGEs have attempted to incorporate logic rules, either predefined or learned, to improve the models' expressiveness\cite{rocktaschel2015injecting}\cite{zhang2019iteratively}.
Efforts have been made to merge KGE and logic rules for enhanced inference, but these attempts have either added logic rules as constraints to KGE's loss functions like done by  KALE \cite{guo2016jointly}, and RUGE \cite{guo2018knowledge}), or by probability-based models which approximate logic inference like MAX-SAT~\cite{qu2019probabilistic,zhang2019can,harsha2020probabilistic}. Moreover, the majority of the approaches need to sample ground rules to tackle the scalability issue, and this limits their ability to effectively process logic rules. Other models try to address these challenges by restricting the logic rules to special kind of FOL, like for example UniKER~\cite{cheng2021uniker}.

\section{Conclusions}
\label{sec:conc}

The PharmKG dataset has been recently proposed as a new biomedical knowledge graph describing interactions between genes, drugs and diseases. Due to the size of the dataset and the multi-relational setting, achieving high performance on this benchmark is very challenging. In this paper, we show how exploiting additional higher-order knowledge in form of First-Order Logic rules, using a novel neuro-symbolic methodology called R2N, improves over the current state-of-the-art on PharmKG. More generally, we show how R2Ns are well suited for contexts where logic rules may enhance the available relational knowledge in a given KG. One key advantage of R2Ns is that the provided rules can also be partially valid, as the model is able to decide how strongly and on which ground atoms to enforce the logic knowledge to correctly predict unknown facts. 
In addition, we provide an ablation study to evaluate the effect of selecting different sets of rules, according to different extraction criteria.

This paper focuses its experimental evaluation on the PharmKG benchmark. However, combining rule mining techniques with R2Ns can be successfully applied to several knowledge graph completion tasks, especially in multi-relational domains, 
where a higher-order logic knowledge can be effective in providing additional information to solve the task.
Therefore, in future work we aim at testing this methodology also on other (potentially very) large KGs, both for the biomedical and other  relational application domains.


\bibliographystyle{IEEEtran}
\bibliography{references.bib}

\end{document}